# Data Fusion of Semantic and Depth Information in the Context of Object Detection

1st Md Abu Yusuf
*Faculty of Computer Science*
*Technische Universität Chemnitz*
Chemnitz, Germany
samba.yusuf@gmail.com

2nd Md Rezaul Karim Khan
*Department of Computer Science*
*Maharishi International University*
IA, USA
mdrezaulkhan@gmail.com

3rd Partha Pratim Saha
*Department of Computer Science*
*Maharishi International University*
IA, USA
sahapratim2@gmail.com

4th Mohammed Mahbubur Rahaman
*Department of Computer Science*
*Maharishi International University*
IA, USA
mahbubur_rahaman@outlook.com

*Abstract*—Considerable study has already been conducted regarding autonomous driving in modern era. An autonomous driving system must be extremely good at detecting objects surrounding the car to ensure safety. In this paper, classification, and estimation of an object's (pedestrian) position (concerning an ego 3D coordinate system) are studied and the distance between the ego vehicle and the object in the context of autonomous driving is measured. To classify the object, faster Region-based Convolution Neural Network (R-CNN) with inception v2 is utilized. First, a network is trained with customized dataset to estimate the reference position of objects as well as the distance from the vehicle. From camera calibration to computing the distance, cutting-edge technologies of computer vision algorithms in a series of processes are applied to generate a 3D reference point of the region of interest. The foremost step in this process is generating a disparity map using the concept of stereo vision.

## I. INTRODUCTION

Artificial intelligence has been developing at a rapid pace in recent times, and this has progressively made it one of the main reason underlying the advancement of autonomous driving technology. While artificial intelligence is helping autonomous driving technology advance significantly, there are still numerous obstacles to overcome, including safety being a particularly important one. It is very crucial to employ deep learning methods to enhance the autonomous driving system's understanding of its surroundings. High-precision object identification is the basis of excellent perceptual ability among them. Actions to prevent incidents could be achieved by using object detection to identify and recognize changes in outside objects [1].

Autonomous driving relies heavily on 3D object detection for the accuracy of their vision system. An object detection technique accumulates and assesses physical attributes (positions, shapes and so on) of an object [2]. However, identifying objects in real-world images can be difficult due to various factors such as lighting, object size and location, object distortions, and issues with imbalance involving both class and objective imbalance. Therefore, an improved Deep Multi-modal 3D detection system is proposed in [3] which utilizes LiDAR and image data. Current detection techniques which employ the sense of heterogeneous data from several sensors are investigated with real and simulated information in [4] to find out the effectiveness of these methods. FPGA-based detection technique consisting of various CMOS cameras are presented in [5] where the proposed system is controlled by Python-based software.

Pedestrians are among the most crucial items to identify in the field of autonomous driving, where 3D object detection is a critical issue. This has been demonstrated that the newly created PointPillars topology works well. In order to make PointPillars a repetitive network, the number of LiDAR frames used in each forward pass is reduced in [6]. In particular, the proposed model only employs 3 frames and recurrent memory instead of 10 LiDAR frames and provides 8% improvement in pedestrian recognition. Another real time detector depending on LiDAR for RGB images is developed in [7] with the help of YOLOv4 which performed accurately and effectively when compared to existing techniques.

In overall, difference in size is a significant barrier in object detection. Feature pyramid networks, that utilize multi-level features obtained as the backbone for top-down upsampling and fusion to gather a set of multi-scale depth characteristics of images, are currently used in many ways to address the issues that arise from a broad range of elements in object detection. An adaptive feature pyramid method is developed with upsampling and fusion designs and validated with experimental results in [8]. Studies associated with the integration of radar data into additional sensors are presented in [9]. The radar data are crucial, particularly when precipitation and weather have an impact on the accuracy of the information. Here, it is important to have some sensors and radar being one of them that are resistant to many types of weather.

In order to quantify the distance between a car and a pedestrian, this study combined depth information with bounding box data for object recognition. However, it must only take into account the pertinent data, regardless of whether it is a disparity map or a bounding box. A very high prediction score is required for the bounding box that has been discovered. It's referred to as the threshold here. Computed the means of all the non-zero disparities inside the enclosing box. This investigation therefore validated the merging of semantic data. Although it is unique, the data fusion method is not autonomous. This is dependent on the stereo camera calibration and accurate image rectification. An algorithm is created in order to accomplish this sort of task.

## II. THEORETICAL BACKGROUND

### A. Convolutional Neural Networks

Within the class of artificial neural networks that have dominated computer vision, CNNs are the most well-established algorithm among numerous deep learning models. It goes without saying that CNN has the ability to advance multiple areas, including image reconstruction, segmentation, and classification, as well as autonomous driving. The input layer receives the input, which is often a multidimensional vector. From there, it distributes the data to the subsequent levels (hidden layers). The hidden layers are going to consider how a stochastic variation inside itself affects or enhances the ultimate result, making judgments depending on the input from the preceding layer. The network develops as a result. Deep learning is a technique that involves stacking numerous hidden layers on top of one another [10].

### B. Faster R-CNN

The latest achievements using RCNNs have propelled significant progress in object detection. To categorize the proposed areas into types of objects or background, CNNs are trained end-to-end using the R-CNN approach. R-CNN functions primarily as a predictor; aside from refinement via bounding box regression, this fails to forecast object limits. The region suggestion module's efficiency determines how accurate it is [11]. Faster R-CNN architecture is shown in Fig. 1. Faster R-CNN and SSD have been utilized in [12] to identify objects, and the assessment demonstrated that the proposed approach may be employed for radar imaging for autonomous applications.

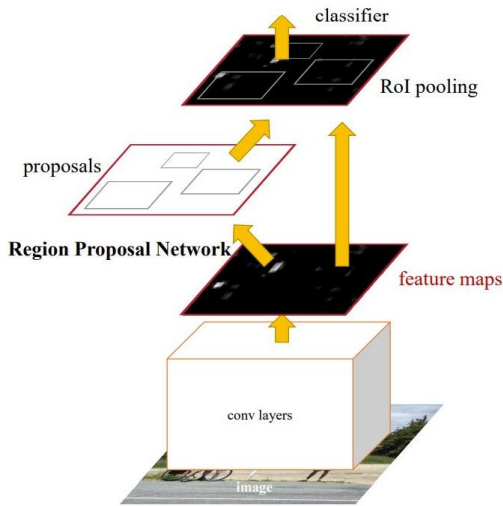

Fig. 1. Architecture of Faster R-CNN [10].

### C. 3D Reconstruction of Scenes

3D reconstruction is a technique used in computer vision that replicates the appearance and geometry of real things. It may be used to find the object's profile and the 3D coordinate of any point on the profile. In many domains, including computer-aided geometric design, graphics and animation, computer vision, medical imaging, computational science, virtual reality, digital media, etc., the 3D reconstruction of objects is a fundamental technique. It is possible to do 3D reconstruction by active or passive techniques. Active techniques use a numerical estimation approach to build a substance's 3D profile based on the provided depth map. However, passive techniques of 3D reconstruction do not alter the rebuilt object; instead, they rely solely on image interpretation and a sensor that detects the illumination reflected or released by the object's exterior to deduce its 3D structure [13].

### D. Stereo Vision and Camera Calibration

One method of estimating depth using two or more cameras is called stereo vision. The associated methods were established over several decades, and the subject matter is wide in the field of computer vision development. Nonetheless, the focus of the study is on methods that are practical to apply in real time. Consequently, a stereo camera made up of two or more cameras serves as the kernel hardware for the stereo vision approach. In addition to correcting lens distortions, camera calibration computes the camera matrix using readings taken in the actual three-dimensional environment. Furthermore, it is also feasible to do 3D reconstruction using the camera matrix. The extrinsic and intrinsic characteristics are used by the calibration process to compute the camera matrix [14].

## III. IMPLEMENTATION AND RESULTS ANALYSIS

The details of system implementation and result analysis are described in this section. All the steps followed to get the output and Local Dynamic Map are shown in Fig. 2. At first, the system takes two RGB images using two pinhole calibrated cameras as input. After that, the images are rectified to bring them in the same plan. Then, the 2D object detection algorithm detects pedestrian(s) in the image if there are any, and the depth map generation algorithm uses two images to generate a depth map. In the next step, the outcome, i.e., data of the two steps, is fused, finding the mean of valid disparity points in the image. Hereafter, the 3d construction algorithm finds the world point of the mean point. In the last step, a visual representation of the world point is shown in an image called a Local Dynamic Map.

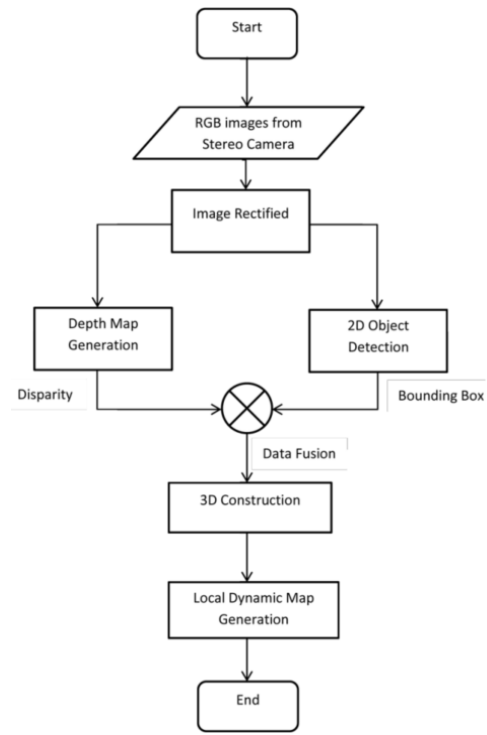

Fig. 2. Steps those are followed to generate final output.

## A. 2D Object Detection

Two important approaches are used in this work to produce a solid solution. Among the techniques is 2D object detection. Using this technique, a deep model is created to identify the 2D item in an image. The research takes into account the best and most effective state-of-the-art architecture since the model may identify the item in real time. The most effective combination utilizes the most recent deep learning methods: Faster RCNN and Inception v2. The architecture-based methodology is well suited to provide output in real time. A few steps have been taken in order to obtain the final model.

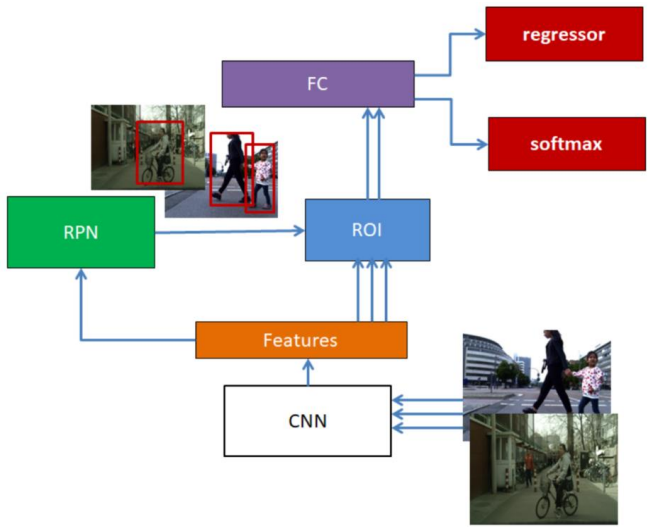

Fig. 3. A schematic overview of 2D Object Detection.

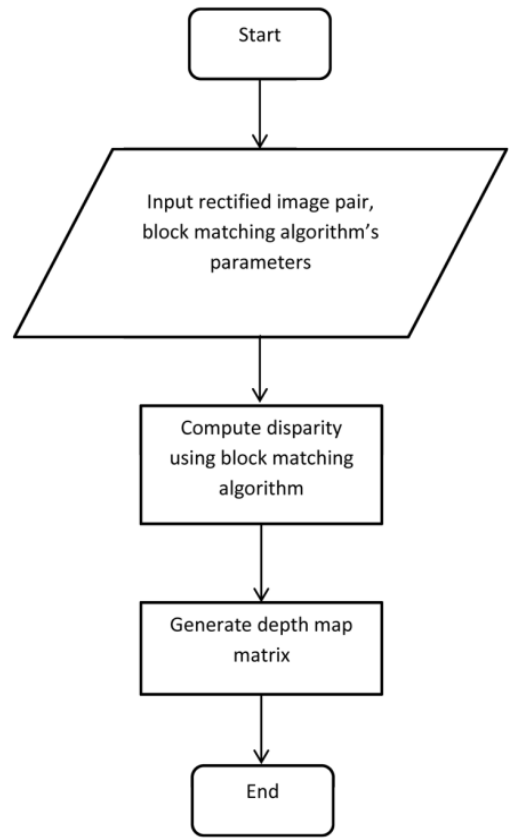

Fig. 4. Depth Map Generation algorithm.

A schematic overview of the architecture of 2D object detection is shown in Fig. 3. The network takes a real-time image as input, then outputs an estimated 2D bounding box around the object. CNN is the backbone for object detection. The CNN, developed with the architecture inception v2. The cost for extracting features is reasonable as the CNN is getting wider than deeper. The deeper network takes more time to process information. The RGB image captured by the camera sensor from the environment passes to the Convolution Neural Network. Then it resizes the input image to 480px with a shorter side and 640px with the longer side, along with three color channels. The feature map would be 30x40x256 dimensions. Feature map only contains detected objects, i.e., pedestrians. RPN (Region Proposal Network) proposes regions or candidate boxes. The region Of Interest (ROI) pooling layer produces a fixed-size feature map. The final Fully Connected (FC) layer is used to check the bounding box of ROI using the regression method and predict the class name of the detected object using SoftMax.

## B. Depth Map Generation

Since the disparity computation is necessary for precise distance measurement, the creation of the depth map is the most important step. As a result, two corrected photos have been passed via a set block of 25 pixels. As a result, a matrix representing the block disparity has been calculated and saved. Algorithm of Depth Map Generation and Depth map of pedestrians are depicted in Fig. 4 and Fig. 5 (a, b).

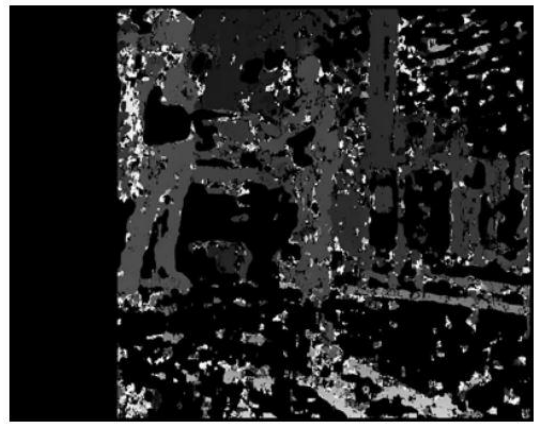

(a)

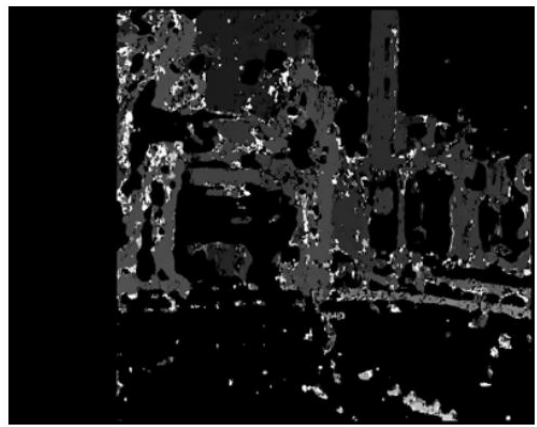

(b)

Fig. 5 (a, b). Depth Map of Pedestrians.

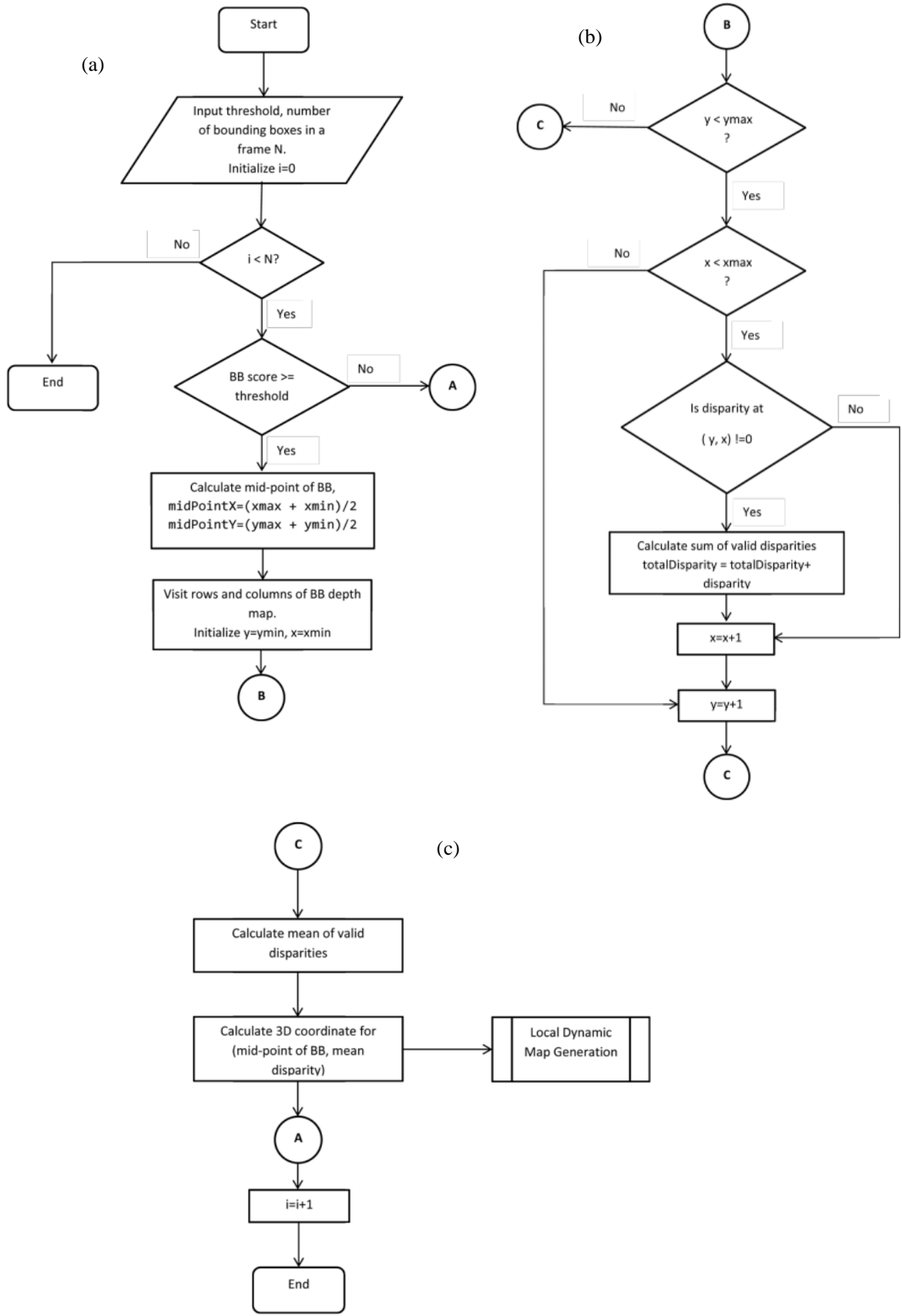

Fig. 6 (a, b, c). Algorithm to calculate mean disparity of each Bounding Box (BB) and construction of world coordinate of corresponding mean disparity at x and y point in the depth map matrix.

## C. Semantic Data Fusion

Data from the two sources are integrated into this step to produce meaningful, accurate, and consistent information that is not possible by a single source is called data fusion. Hence, two data sources are the bounding box and the disparity map.

The data which are valid and meaningful to produce accurate output is called semantic data. For example, this study treats the non-zero value of disparity with in the bounding box border as semantic data. Algorithm to calculate mean disparity is shown in Fig. 6 (a, b, c). Circled A and B from Fig. 6 (a) and Circled C from Fig. 6 (b) denoted that steps are continued to the following Fig. 6 (b) and Fig. 6 (c), simultaneously.

*D. Creation of Local Dynamic Map*

The localization of pedestrian respect to the car within 6 meters boundary has been shown for every frame on the generated image. It's live. To do this task, algorithms which are followed is depicted in Fig. 7. To see the live update of objects, it is necessary to create a video that consists of frames. And every frame is an image. Here it is called base image. Each base image has height and width. Set offset at the beginning of the algorithm to draw circles in the upper left, right corner, and middle of the image. A homography matrix is created. Homography matrix is used to map between two planes, base image, and world coordinate, i.e., mean point of pedestrian. After mapping, the algorithm checks whether the transformed world coordinates inside or not inside of base image.

*E. Result Analysis*

Two pedestrians' position in a frame when they are in a certain location A is shown in Fig. 8. The frame shows the position of two pedestrians in front of a car in the distance of 3.26 meter and 3.87 meter respectively. The blue circle act as a front of the car. The green and yellow circles indicate the view range of side and front of Local Dynamic Map. Here max view range in any side is 2.5 meters and in the front is 6 meters in respect to car.

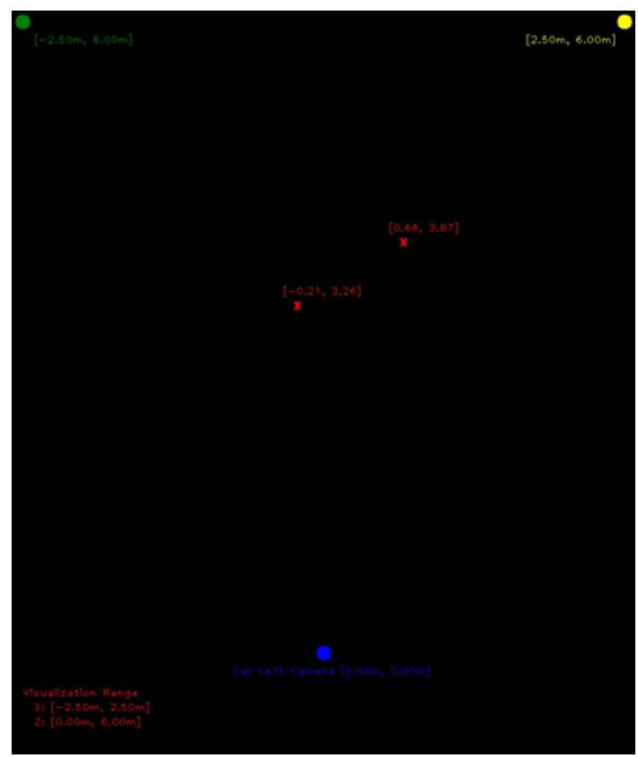

Fig. 8. Position of two pedestrians in certain location A.

On the other hand, two pedestrians' position in a frame when they are in a certain location B is depicted in Fig. 9. This second frame shows the position of two pedestrians in front of a car in the distance of 2.93 meter and 3.74 meter respectively. The blue circle act as a front of a car. The green and yellow circles indicate the view range of side and front of Local Dynamic Map. Here max view range in any side is 2.5 meters and in the front is 6 meters in respect to car.

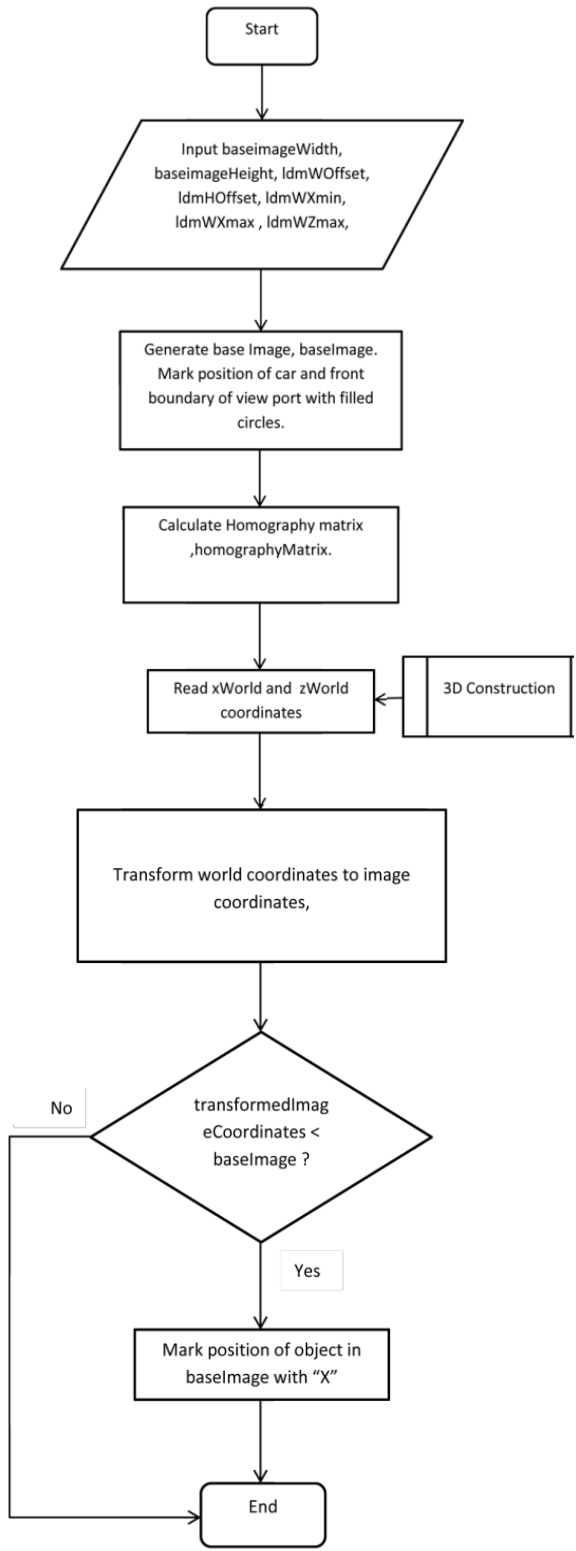

Fig. 7. Algorithm to create Local Dynamic Map.

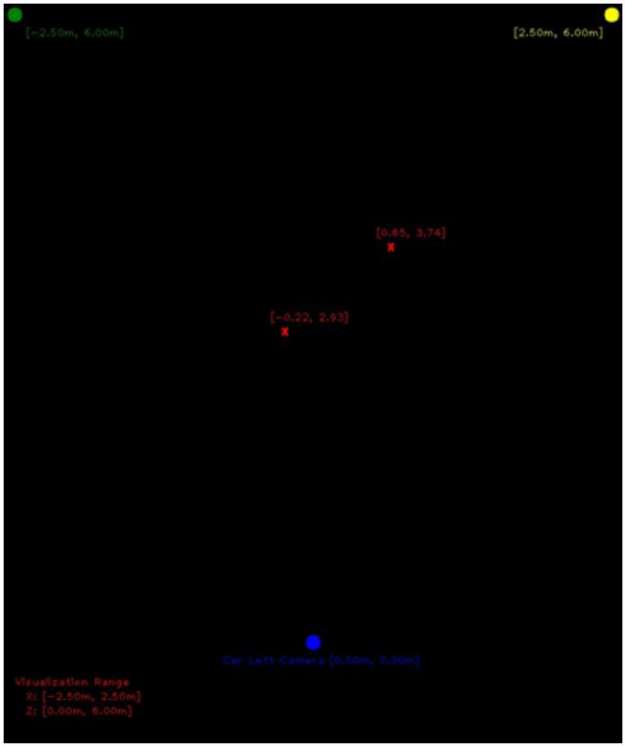

Fig. 9. Position of two pedestrians in certain location B.

## IV. Conclusion

In conclusion, it can be said that the developed algorithm was able to achieve its goal. Even it was implemented successfully and was able to show the position of the objects respect to the car position in a live screen, Local Dynamic Map. The algorithm started with camera calibration. Every time the calibration is not required, because the parameters were saved in a file, until there are some changes in the hardware and software. It makes the process faster. It is very much necessary to get the information in low latency in the Local Dynamic Map while the car is running in the highway. The faster pedestrian detection algorithm, faster computation, i.e. consider a block of pixels instead of going through every pixel, to generate disparity map and pick only valid value of disparities and then calculate world coordinate of a mean value ensured a standard LDM. The study started with a motivation to reduce accidents on the road. The motivation will be fulfilled if this developed algorithm is integrated as a part of autonomous driving system.

Like other algorithms, it has limitations too. For example, it did not produce an accurate object position when it was closer to the car or camera and far away, i.e. 6 meters, from the car or camera. In addition, the algorithm was not tested at night, in foggy environments, in rainy and snowing weather, or on hill tracks. Therefore, there are options to overcome such types of distance, environmental, and location-based challenges.